\def\year{2021}\relax
\documentclass[letterpaper]{article} 

\usepackage{amsmath,amsfonts,bm}

\def\eqref#1{equation~\ref{#1}}

\def\1{\bm{1}}

\DeclareMathAlphabet{\mathsfit}{\encodingdefault}{\sfdefault}{m}{sl}
\SetMathAlphabet{\mathsfit}{bold}{\encodingdefault}{\sfdefault}{bx}{n}

\def\sX{{\mathbb{X}}}

\usepackage{microtype}
\usepackage{color}
\usepackage[caption=false]{subfig}

\usepackage{booktabs} 
\usepackage[flushleft]{threeparttable}
\usepackage{tabularx}
\usepackage{multirow}

\newcommand{\tabincell}[2]{\begin{tabular}{@{}#1@{}}#2\end{tabular}}

\usepackage{aaai21}  
\usepackage{times}  
\usepackage{helvet} 
\usepackage{courier}  
\usepackage[hyphens]{url}  
\usepackage{graphicx} 
\usepackage{graphicx}  
\usepackage[square,sort,comma,numbers]{natbib}  
\usepackage{caption} 
\title{Generalize a Small Pre-trained Model to Arbitrarily Large TSP Instances}
\author {

        Zhang-Hua Fu \textsuperscript{\rm 1,2},
        Kai-Bin Qiu \textsuperscript{\rm 2},
        Hongyuan Zha \textsuperscript{\rm 1,3 \thanks{Corresponding author.} } \\
}
\affiliations {
    \textsuperscript{\rm 1} Shenzhen Institute of Artificial Intelligence and Robotics for Society, Shenzhen, China \\
    \textsuperscript{\rm 2} Institute of Robotics and Intelligent Manufacturing, The Chinese University of Hong Kong, Shenzhen, China \\
    \textsuperscript{\rm 3} School of Data Science, The Chinese University of Hong Kong, Shenzhen, China \\
    fuzhanghua@cuhk.edu.cn, 220019002@link.cuhk.edu.cn, zhahy@cuhk.edu.cn
}
\begin{document}

\maketitle

\begin{abstract}
For the traveling salesman problem (TSP), the existing supervised learning based algorithms suffer seriously from the lack of generalization ability. To overcome this drawback, this paper tries to train (in supervised manner) a small-scale model, which could be repetitively used to build heat maps for TSP instances of arbitrarily large size, based on a series of techniques such as graph sampling, graph converting and heat maps merging. Furthermore, the heat maps are fed into a reinforcement learning approach (Monte Carlo tree search), to guide the search of high-quality solutions. Experimental results based on a large number of instances (with up to 10,000 vertices) show that, this new approach clearly outperforms the existing machine learning based TSP algorithms, and significantly improves the generalization ability of the trained model.
\end{abstract}

\section{Introduction}
\label{Introduction}

The travelling salesman problem (TSP) is a well-known combinatorial optimization problem with various real-life applications, such as transportation, robots routing, biology, circuit design. Given $n$ cities as well as the distance $d_{ij}$ between each pair of cities $i$ and $j$, the TSP aims to find a cheapest tour which starts from a beginning city (arbitrarily chosen), visits each city exactly once, and finally returns to the beginning city. This problem is NP-hard, thus being extremely difficult from the viewpoint of theoretical computer science.

Due to its importance in both theory and practice, many algorithms have been developed, mostly based on traditional operations research (OR) methods. Among the existing TSP algorithms, the best exact solver Concorde \citep{applegate2009certification} succeeded in demonstrating optimality of an Euclidean TSP instance with 85,900 cities, while the leading heuristics \citep{helsgaun2017extension} and \citep{taillard2019popmusic} are capable of obtaining near-optimal solutions for instances with millions of cities. However, these algorithms are very complicated, which consist of many hand-crafted rules and heavily rely on expert knowledge, thus being difficult to generalize to other combinatorial optimization problems.

To overcome those limitations, recent years have seen a number of machine learning (ML) based algorithms being proposed for the TSP (briefly reviewed in the next section), which attempt to automate the search process by learning mechanisms. This type of methods do not rely heavily on expert knowledge, can be easily generalized to various combinatorial optimization problems, thus become a promising research direction.

For the TSP, existing ML based algorithms can be roughly classified into two categories, i.e., (1) supervised learning (SL) algorithms which attempt to discover common patterns supervised by pre-computed TSP solutions. (2) reinforcement learning (RL) algorithms which try to learn during the interaction with the environment (without pre-computed solutions).

Once well trained, SL models are able to provide useful information that significantly speeds up the search of high-quality TSP solutions. However, the performance of a pre-trained model of fixed size may decrease drastically while tackling TSP instances of different sizes, since the distributions of the training instances are very different from the test instances. On the other hand, training SL models generally requires a large number of pre-computed optimal (at least high-quality) TSP solutions, being unaffordable for large-scale TSP instances. These drawbacks seriously limit the usage of SL on large-scale TSP instances.

However, we believe the idea of discovering common patterns in a supervised manner is valuable. If we can train a small-scale SL model within reasonable time and find a way to smoothly generalize it to large-scale cases (without pre-computing a large number of solutions again), it is hopeful to inherit the advantages of SL while avoiding its drawbacks. Motivated by this idea, we develop a series of techniques, in order to improve the generalization ability of the model trained by SL. Furthermore, we combine SL and RL to form a hybrid algorithm, which performs favorably with respect to the existing ML based TSP algorithms. Overall, the main contributions are summarized as follows.

\begin{itemize}

  \item \textbf{Methodologies}: At first, we train a small-scale (with size $m$) model by supervised learning, based on a graph convolutional residual network with attention mechanism (Att-GCRN). Once well trained, given a TSP instance with $m$ vertices, the model is able to build a heat map over the edges. Then, we try to smoothly generalize this model to handle large instances. For this purpose, given a large-scale TSP instance, we repeatedly use a graph sampling method to extract a sub-graph with exactly $m$ vertices, then convert it to a standard TSP instance, and call the pre-trained model to build a sub heat map. Finally, all the sub heat maps are merged together, to get a complete heat map over the original graph. Although the Att-GCRN is somewhat similar to the network in \citep{joshi2019efficient}, to our best knowledge, the graph sampling, graph converting and heat maps merging techniques are firstly developed for the TSP in this paper, which significantly improve the generalization ability of the trained model.

      Furthermore, based on the merged heat map, we use a RL based approach, i.e., Monte Carlo tree search (MCTS), to search high-quality solutions. To our best knowledge, there are two existing works \citep{shimomura2016} and \citep{xing2020a} which also use MCTS to solve the TSP. However, they are both constructive approaches, where each state is a partial TSP tour, and each action adds a city to increase the partial tour. By contrast, our MCTS method is a conversion based approach, where each state is a complete tour, and each action converts the current state to a new complete tour. Therefore, our method is very different from the existing MCTS algorithms.

  \item \textbf{Results}: We carry out experiments on a large number of TSP instances with up to 10,000 cities (one order of magnitude larger than the instances used to evaluate the existing ML algorithms). On all the data sets, our new algorithm is able to obtain optimal or near-optimal solutions within reasonable time, clearly outperforming all the existing learning based algorithms.
\end{itemize}

\section{Related works}
\label{RelatedWork}

In this section, we briefly review the existing ML based algorithms on the TSP, and then extend to several other highly related problems. Non-learned methods are omitted, interested readers please find in \citep{applegate2009certification}, \citep{rego2011traveling}, \citep{helsgaun2017extension} and \citep{taillard2019popmusic} for an overlook of the leading TSP algorithms.

The idea of applying ML to solve the TSP dated back to several decades ago \citep{hopfield1985neural}, but becomes a hot and promising topic only in recent years. A number of ML based TSP algorithms have been developed, which can be classified into two categories.


\textbf{Supervised learning (SL) methods:} \cite{vinyals2015pointer} introduced a pointer network which consists of an encoder and a decoder, both using recurrent neural network (RNN). The encoder parses each TSP city into an embedding, and then the decoder uses an attention model to predict the probability distribution over the candidate (unvisited) cities. \cite{nowak2017note} proposed a supervised approach, which trains a graph neural network (GNN) to predict an adjacency matrix (heat map) over the cities, and then attempts to convert the adjacency matrix to a feasible TSP tour by beam search (OR based method). \cite{joshi2019efficient} followed this framework, but chose deep graph convolutional networks (GCN) to build heat map, and then constructed tours via highly parallelized beam search. \cite{xing2020a} trained a graph neural network (GNN) to capture the local and global
graph structure, based on which they used a MCTS procedure to construct TSP tours. These SL based methods require a large number of pre-computed TSP solutions, thus being difficult to directly generalize to large-scale instances.

\textbf{Reinforcement learning (RL) methods:} To overcome the drawback of SL, several groups chose RL instead of SL. For example, \cite{bello2016neural} implemented an actor-critic RL architecture, which uses the tour length as a reward, to guide the search towards promising area. \cite{khalil2017learning} proposed a framework which maintains a partial tour and repeatedly calls a RL model to select the most relevant city to add to the partial tour, until forming a complete TSP tour. \cite{emami2018learning} also implemented an actor-critic neural network, and chose Sinkhorn policy gradient to learn policies by approximating a double stochastic matrix. Concurrently, \citep{deudon2018learning}, \citep{kool2018attention} both proposed a graph attention network (GAN), which incorporates attention mechanism with RL to auto-regressively improve the quality of the obtained solution.


In addition to the works focused on the classic TSP, there are several ML based methods recently proposed for other related problems, such as the decision TSP \citep{prates2019learning}, the multiple TSP \citep{kaempfer2018learning}, and the vehicle routing problem \citep{nazari2018reinforcement}, \citep{chen2019learning} and \citep{lu2020a}, etc. For an overall survey, please refer to \citep{bengio2018machine} and \citep{guo2019solving}.

\section{Methods}
\label{Methods}

\subsection{Preliminaries}
\label{Preliminaries}

In this paper, we focus on the two-dimensional Euclidean TSP, which is formulated as an undirected graph $G(V, E)$, where $V$ (with $|V|=n$) denotes the set of vertices (each vertex corresponds to a city), and $E$ denotes the set of edges. Without loss of generality, assume all the vertices are distributed within a two dimensional unit square, i.e., for each vertex $i \in V$, its coordinates $x_i$ and $y_i$ both belong to $[0,1]$, and the distance $d_{ij}$ is defined as the Euclidean distance between vertices $i$ and $j$. Furthermore, corresponding to graph $G$, its heat map is defined as a $n \times n$ matrix $\bm{P}$, whose element $P_{ij} \in [0,1]$ denotes the probability of edge $(i,j)$ belonging to the optimal TSP solution.

As a preliminary step, we at first train ({off-line learning}) a graph convolutional residual neural network with attention mechanisms (denoted by Att-GCRN for short, whose architecture is described in the full version of this paper \footnote{https://github.com/Spider-scnu/TSP}), with fixed input size $m$ (a parameter). To train the model, 990,000 TSP instances with $m$ vertices are randomly generated as the train set, and the solutions produced by the exact solver Concorde \citep{applegate2006concorde} are used as the ground-truth solutions. Once well trained, given a new TSP instance with $m$ vertices (randomly distributed within an unit square), the model is able to build a heat map, which estimates the probability $P_{ij}$ of each edge $(i,j)$ belonging to the optimal solution.

\begin{figure}[!htbp]
  \centering
  \includegraphics[width=0.5\textwidth]{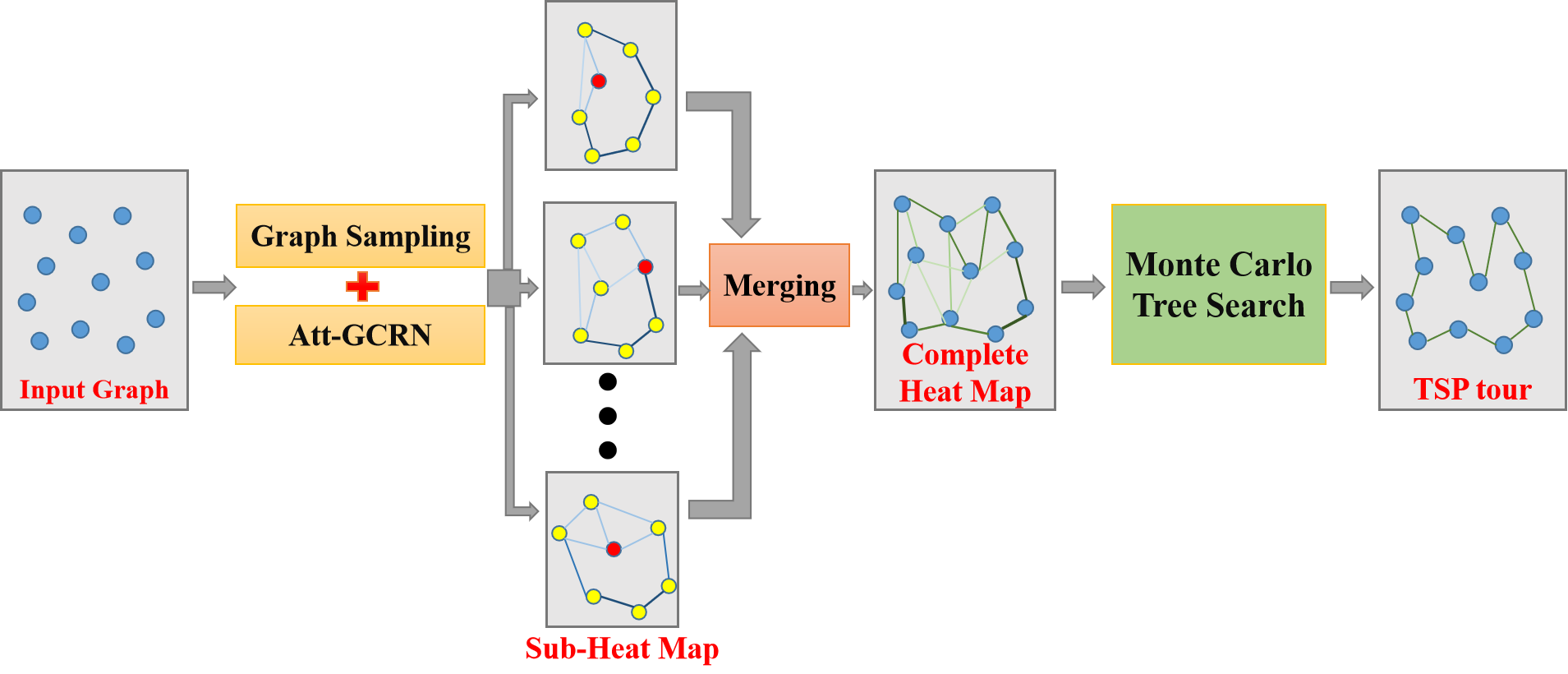}
  \caption{Pipeline of the proposed approach}
  \label{Fig.framework}
\end{figure}

\subsection{Pipeline}
\label{Pipeline}

Given a TSP instance of arbitrarily large size, the pipeline for solving this instance is shown in Fig. \ref{Fig.framework}, which consists of three main steps. Respectively, the first step (off-line learning) uses a graph sampling method to extract from the original graph a number of sub-graphs (each exactly consists of $m$ vertices), and then uses the pre-trained Att-GCRN model to build a sub heat map corresponding to each sub-graph. After that, the second step tries to merge all the sub heat maps into a complete heat map (corresponding to the original graph). Finally, the third step uses a reinforcement learning method ({online learning}), i.e., Monte Carlo tree search (MCTS), to search high-quality TSP solutions, guided by the information stored in the merged heat map.

\subsection{Building and merging heat maps}
\label{Heatmap}

\begin{figure}[!htbp]
  \centering
  \includegraphics[width=0.5\textwidth]{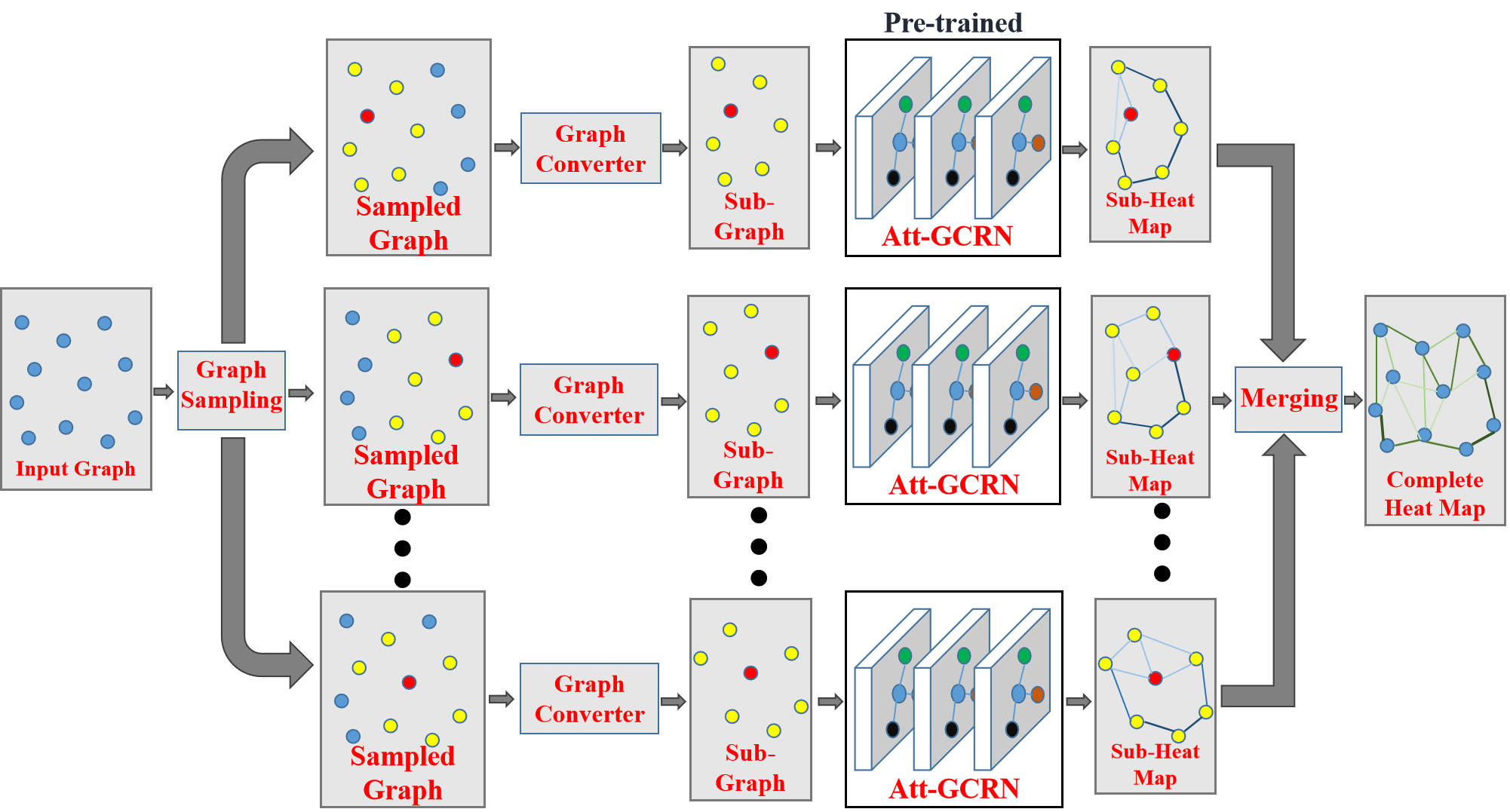}
  \caption{Method for building and merging heat maps}
  \label{Fig.Heatmap}
\end{figure}

The pre-trained model is able to build a heat map of a TSP instance with $m$ vertices. However, it can not be directly used to handle instances of different size. To deal with this issue, an optional approach is to train a series of models with different sizes, like the choice of \citep{joshi2019efficient}. Unfortunately, this approach seems unreasonable for very large TSP instances, since the supervised learning process requires a large number of pre-computed optimal (at least high-quality) solutions, being unaffordable for large-scale TSP instances. To avoid repetitively training models, in this paper we develop a series of techniques (illustrated in Fig. \ref{Fig.Heatmap} and described as follows), to extend the predication ability of the fix-sized model to arbitrarily large TSP instances.

\subsubsection{Graph sampling}
\label{GraphSamling}

The graph sampling method is used to extract a number of sub-graphs (each with $m$ vertices) from the original graph $G$. To do this, for each vertex $i \in V$ or each edge $(i,j) \in E$, let $O_i$ or $O_{ij}$ (initialized to 0) respectively denote the times that vertex $i$ (or edge $(i,j)$) belongs to an extracted sub-graph. Then, at each iteration, we choose the vertex $i$ with the minimal value of $O_i$ (randomly choose one if there are multiple such vertices) as the clustering center, and use the k-nearest neighbors algorithm \citep{dudani1976distance} to extract a sub-graph $G'$ which consists of exactly $m$ vertices (including the clustering center). Then, for each vertex $i$ or each edge $(i,j)$ belonging to $G'$, let $O_i \leftarrow O_i+1$, $O_{ij} \leftarrow O_{ij}+1$.

Above process is repeated, until the minimal value of $O_i$ reaches a lower bound $\omega$ (a pre-defined parameter). Notice that the extracted sub-graphs may overlap, i.e., any vertex or edge may belong to different sub-graphs.

\subsubsection{Graph converting}
\label{GraphConverting}

For each instance of the train set, all the vertices are distributed randomly within an unit square. To make sure the extracted sub-graph $G'$ also meets this distribution, we should convert it to a new graph $G''$. For this purpose, let $x^{min} = \min \limits_{i\in G'} x_i$, $x^{max} = \max \limits_{i\in G'} x_i$, $y^{min} = \min \limits_{i\in G'} y_i$, $y^{max} = \max \limits_{i\in G'} y_i$ respectively denote the minimal, maximal value of the horizonal and vertical coordinates among all the $m$ vertices of $G'$, and let $s = \frac{1}{\max (x^{max} - x^{min}, y^{max} - y^{min})}$ be an amplification factor. Then, for each vertex $i \in G'$, we convert its coordinates $(x_i,y_i)$ to new coordinates $(x^{new}_i,y^{new}_i)$:

\begin{equation}
\label{scale}
    \begin{aligned}
    x^{new}_i &\leftarrow s \times (x_i - x^{min}), \\
    y^{new}_i &\leftarrow s \times (y_i - y^{min}).
    \end{aligned}
\end{equation}

After that, sub-graph $G'$ is converted to a new graph $G''$.

\subsubsection{Building sub heat maps}
\label{BuildSubHeatMaps}

For each converted sub-graph $G''$, the coordinates of the $m$ vertices are fed into the pre-trained Att-GCRN model, to build a sub heat map over $G''$.

\subsubsection{Merging sub heat maps}
\label{MergeSubHeatMaps}

Above two steps are repeated, thus we can obtain a number (denoted by $I$) of sub heat maps. Finally, we try to merge them into a complete heat map. To do this, for each edge $(i,j)$ of the original graph $G$, we estimate its probability $P_{ij}$ of belonging to the optimal TSP solution as follows.

\begin{equation}
    \label{equ:updateprob}
    \begin{aligned}
        P_{ij} &= \frac{1}{O_{ij}} \times \sum \limits_{l=1}^{I} P_{ij}^{''}(l).
    \end{aligned}
\end{equation}

where $P_{ij}^{''}(l)$ denotes the probability of edge $(i,j)$ (after conversion) belonging to the optimal solution of the $l$th converted sub-graph $G''$ .

After merging all the sub heat maps, we obtain a complete heat map over the original graph $G$. Then, all the edges with $P_{ij} < 10^{-4}$ are marked as unpromising edges, which are eliminated directly to reduce the search space.

\newpage
\subsection{Reinforcement learning for solutions optimization}
\label{RLForOptimization}

Based on the heat map obtained above, we develop a reinforcement learning based approach to search high-quality solutions. The search process is considered as a Markov Decision Process (MDP), which starts from an initial state $\bm{\pi}$, and iteratively applies an action $\bm{a}$ to reach a new state $\bm{\pi}^*$. The details are described as follows.

\subsubsection{States and actions}
\label{StateAndAction}
In our implementation, each state corresponds to a complete TSP solution, i.e., a permutation $\bm{\pi} =(\pi_1,\pi_2,\ldots,\pi_n)$ of all the vertices. Each action $\bm{a}$ is a transformation which converts a given state $\bm{\pi}$ to a new state $\bm{\pi}^*$. Since each TSP solution consists of a subset of $n$ edges, any action could be viewed as a $k$-opt ($2 \le k \le n$) transformation, which deletes $k$ edges at first, and then adds $k$ different edges to form a new tour.

Obviously, each action can be represented as a series of $2k$ sub-decisions ($k$ edges to delete and $k$ edges to add). This representation method is straightforward, but seems a bit redundant, since the deleted edges and added edges are highly relevant, while arbitrarily deleting $k$ edges and adding $k$ edges may result in an unfeasible solution. To overcome this drawback, we develop a compact method to represent an action, which consists of only $k$ sub-decisions. Formally, an action can be represented as $\bm{a}=(a_1,b_1,a_2,b_2,\dots,a_k,b_k,a_{k+1})$, where $k$ is a variable and the final vertex must coincide with the first vertex, i.e. $a_{k+1}=a_1$. Each action corresponds to a $k$-opt transformation, which deletes $k$ edges, i.e., $(a_i,b_i), 1\le i \le k$, and adds $k$ edges, i.e., $(b_i,a_{i+1}), 1\le i \le k$, to reach a new state. Notice that not all these elements are optional. Once $a_i$ is known, $b_i$ can be uniquely determined without any optional choice (explained and exemplified in the full version of this paper). Therefore, to determine an action we should only decide a series of $k$ sub-decisions, i.e., the $k$ vertices $a_i, 1\le i \le k$. Additionally, an action involving an unpromising edge $(b_i,a_{i+1})$, i.e., $P_{b_ia_{i+1}} < 10^{-4}$, is marked as an unpromising action and eliminated directly.

Intuitively, this compact representation method brings advantages in two-folds: (1) fewer (only $k$, not $2k$) sub-decisions need to be made; (2) the resulting states are necessarily feasible solutions.

Let $L(\bm{\pi})$ denote the tour length corresponding to state $\bm{\pi}$, then corresponding to each action $\bm{a}=(a_1,b_1,a_2,b_2,\dots,a_k,b_k,a_{k+1})$ which converts $\bm{\pi}$ to a new state $\bm{\pi}^*$, the difference $\Delta(\bm{\pi},\bm{\pi}^*)=L(\bm{\pi}^*)-L(\bm{\pi})$ could be calculated as follows:

\begin{equation} \label{delta}
\Delta(\bm{\pi},\bm{\pi}^*)=\sum_{i=1}^{k}d_{b_ia_{i+1}}-\sum_{i=1}^k d_{a_ib_i}.
\end{equation}

If $\Delta(\bm{\pi},\bm{\pi}^*) < 0$, $\bm{\pi}^*$ is better (with shorter tour length) than $\bm{\pi}$.

\subsubsection{State initialization}
\label{Initialize}

For state initialization, we choose a constructive procedure, which starts from an arbitrarily chosen begin vertex $\pi_1$, iteratively selects a vertex $\pi_i, 2 \le i \le n$ among the candidate (unvisited) vertices and adds it to the end of the partial tour, until forming a complete tour $\bm{\pi} =(\pi_1,\pi_2,\ldots,\pi_n)$. More precisely, at the $i$th iteration, if there are more than one candidate vertices, each candidate vertex $j$ is chosen with a probability proportional to $\exp (P_{\pi_ij})$, while all the candidate vertices share a total probability of 1.

\subsubsection{Enumerating within small neighborhood}
\label{Enumerate}

To maintain the generalization ability of our approach, we avoid to use complex hand-crafted rules, such as the $\alpha$-nearness criterion in \citep{helsgaun2000effective} and the POPMUSIC strategy in \citep{taillard2019popmusic}, which have proven to be highly effective on the TSP, but heavily depend on expert knowledge. Instead, starting from a new state, we at first use a straightforward method to search within a small neighborhood. More precisely, the method examines one by one the promising actions with $k=2$, and iteratively applies the first-met improving action which leads to a better state, until no improving action with $k=2$ is found. This simple method is able to efficiently and robustly converge to a local optimal state.

\subsubsection{Targeted sampling within enlarged neighborhood}
\label{MCTS}

Once no improving action is found within the small neighborhood, we switch to an enlarged neighborhood which consists of the actions with $k>2$. Unfortunately, there are generally a huge number of actions within the enlarged neighborhood (even after eliminating the unpromising ones), being impossible to enumerate them one by one. Therefore, we choose to sample a subset of promising actions (guided by RL) and iteratively select an action to apply, to reach a new state.

\begin{figure}[ht]
	\centering	
    \includegraphics[width=1\linewidth]{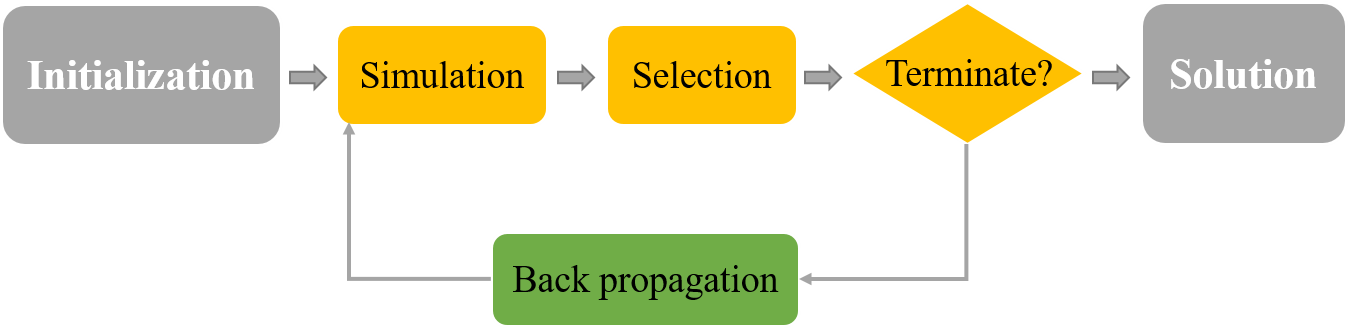}
	\caption{Procedure of the Monte Carlo tree search}
	\label{fig:MCTS}
\end{figure}

Following this idea, we choose the Monte Carlo tree search (MCTS) as our learning framework. Inspired by the works in \citep{coulom2006efficient}, \citep{browne2012survey}, \citep{silver2016mastering} and \citep{silver2017mastering}, our MCTS procedure (outlined in Fig. \ref{fig:MCTS}) consists of four steps, i.e., (1) Initialization, (2) Simulation, (3) Selection, and (4) Back-propagation, which are respectively designed as follows.

\textbf{Initialization:} We define two $n \times n$ symmetric matrices, i.e., a weight matrix $\bm{W}$ whose element $W_{ij}$ (initialized to $100\times P_{ij}$) controls the probability of choosing vertex $j$ after vertex $i$, and an access matrix $\bm{Q}$ whose element $Q_{ij}$ (initialized to 0) records the times that edge $(i,j)$ is chosen during simulations. Additionally, a variable $M$ (initialized to 0) is used to record the total number of actions already simulated. Note that this initialization step should be executed only once at the beginning of the whole process of MDP.

\textbf{Simulation:} Given a state $\bm{\pi}$, we use the simulation process to probabilistically generate a number of actions. As explained before, each action is represented as $\bm{a}=(a_1,b_1,a_2,b_2,\dots,a_k,b_k,a_{k+1})$, containing a series of sub-decisions $a_i, 1\le i \le k$ ($k$ is also a variable, and $a_{k+1}=a_1$), while $b_i$ could be determined uniquely once $a_i$ is known. Once $b_i$ is determined, for each edge $(b_i,j), j\ne b_i$, we use the following formula to estimate its potential $Z_{b_ij}$ (the higher the value of $Z_{b_ij}$, the larger the opportunity of edge $(b_i,j)$ to be chosen):

\begin{equation} \label{EqZij}
Z_{b_ij}=\frac{W_{b_ij}}{\Omega_{b_i}}\ +\alpha \sqrt{\cfrac{\ln\ (M+1)}{Q_{b_ij}+1}}.
\end{equation}

Where $\Omega_{b_i}=\frac{\sum_{j\ne b_i} W_{b_ij}}{\sum_{j \ne b_i} 1}$ denotes the averaged $W_{b_ij}$ value of all the edges relative to vertex $b_i$. In this formula, the left part $\frac{W_{b_ij}}{\Omega_{b_i}}$ emphasizes the importance of the edges with high $W_{b_ij}$ values (to enhance the intensification feature), while the right part $\sqrt{\cfrac{\ln\ (M+1)}{Q_{b_ij}+1}}$ prefers the rarely examined edges (to enhance the diversification feature). $\alpha$ is a parameter used to achieve a balance between intensification and diversification, and the term "+1" is used to avoid a minus numerator or a zero denominator.

To make the sub-decisions sequently, we at first choose $a_1$ randomly, and determine $b_1$ subsequently. Recursively, once $a_i$ and $b_i$ are known, $a_{i+1}$ is decided as follows: (1) if closing the loop (connecting $b_i$ to $a_1$) would lead to an improving action, or $i \ge 10$, let $a_{i+1}=a_1$. (2) otherwise, consider the vertices with $W_{b_ij} \ge 1$ as candidate vertices, forming a set $\sX$ (excluding $a_1$ and the vertex already connected to $b_i$). Then, among $\sX$ each vertex $j$ is selected as $a_{i+1}$ with probability $P_j$, which is determined as follows:

\begin{equation} \label{EqPj}
P_{j}=\cfrac{Z_{b_i j}}{\sum_{l\in \sX}Z_{b_i l}}.
\end{equation}

Once $a_{i+1}=a_1$, we close the loop to obtain an action.

Similarly, more actions are generated (forming a sampling pool), until meeting an improving action which leads to a better state, or the number of actions reaches its upper bound (controlled by a parameter $H$).

\textbf{Selection:} During above simulation process, if an improving action is met, it is selected and applied to the current state $\bm{\pi}$, to get a new state $\bm{\pi}^{new}$. Otherwise, if no such action exists in the sampling pool, it seems difficult to gain improvement within the current search area. Then, the MDP jumps to a random state (using the state initialization method described above), which serves as a new starting state.

\textbf{Back-propagation:} The value of $M$ as well as the elements of matrices $\bm{W}$ and $\bm{Q}$ are updated by back propagation as follows. At first, whenever an action is examined, $M$ is increased by 1. Then, for each edge $(b_i,a_{i+1})$ which appears in an examined action, let $Q_{b_ia_{i+1}}$ increase by 1. Finally, whenever a state $\bm{\pi}$ is converted to a better state $\bm{\pi}^{new}$ by applying action $\bm{a}=(a_1,b_1,a_2,b_2,\dots,a_k,b_k,a_{k+1})$, for each edge $(b_i,a_{i+1}), 1\le i \le k$, let:

\begin{equation} \label{EqWij}
W_{b_ia_{i+1}} \leftarrow W_{b_ia_{i+1}} + \beta  \left[\exp\left(\cfrac{L(\bm{\pi})-L(\bm{\pi}^{new})}{L(\bm{\pi})}\right)-1\right].
\end{equation}

Where $\beta$ is a parameter used to control the increasing rate of $W_{b_ia_{i+1}}$. Notice that we update $W_{b_ia_{i+1}}$ only when meeting a better state, since we want to avoid wrong estimations (even in a bad action which leads to a worse state, there may exist some good edges $(b_i,a_{i+1})$). With this back-propagation process, the weight of the good edges would be increased to enhance its opportunity of being selected, thus the sampling process would be more and more targeted.

$\bm{W}$ and $\bm{Q}$ are symmetric matrices, thus let $W_{a_{i+1}b_i}=W_{b_ia_{i+1}}$ and $Q_{a_{i+1}b_i}=Q_{b_ia_{i+1}}$ always.

\subsubsection{Termination condition}
\label{Termination}

The MCTS iterates through the simulation, selection and back-propagation steps, until no improving action exists among the sampling pool. Then, the MDP jumps to a new state, and launches a new round of search within small and enlarged neighborhood again. This process is repeated, until the allowed time (controlled by a parameter $T$) has been elapsed. Then, the best found state is returned as the final solution.

\section{Experiments}
\label{Experiments}

To evaluate the performance of our method, we program the algorithm for building heat maps in Python, and program the MCTS algorithm in C++ language \footnote{Publicly available at https://github.com/Spider-scnu/TSP.}. Then, we carry out experiments on a large number of TSP instances, and make comparisons with eight newest learning based baselines, as well as three strong non-learning algorithms (the programming and training details about the baselines are given in the full version of this paper). Notice that, for the baselines, we just directly download and rerun the source codes, based on the pre-trained models (only for learning based baselines) which are publicly available. To ensure fair comparisons, all the learning based baselines as well as our new algorithm are uniformly executed on one GTX 1080 Ti GPU (to fully utilize the computing resources, as many instances as possible are executed in parallel). For the three non-learning algorithms, their source codes currently do not support running on GPU, thus we re-run them on one Intel(R) Xeon(R) Gold 5118 CPU @ 2.30GHz (with 8 cores), and list the results just for indicative purposes. Notice that our method is a learning based algorithm, thus we do not aim to strictly outperform the non-learning algorithms.

\subsection{Data sets}
\label{SubsecInstance}

We use two data sets: (1) \textbf{Set 1} \footnote{Downloaded from \url{https://drive.google.com/file/d/1-5W-S5e7CKsJ9uY9uVXIyxgbcZZNYBrp/view}.}, which is divided into three subsets, each containing 10,000 automatically generated 2D-Euclidean TSP instances, respectively with $n=20, 50, 100$. This data set is widely used by the existing learning based algorithms. (2) \textbf{Set 2}, following the same rules, we newly generate 400 larger instances, i.e., 128 instances respectively with $n=200, 500, 1000$, and 16 instances with $n=10000$.

\subsection{Parameters}
\label{SubsecParameter}

As described before, our method relies on six hyper parameters ($m$, $\omega$, $\alpha$, $\beta$, $H$ and $T$). For parameter $m$ which controls the size of the pre-trained model, we set $m=20$ for the small instances of data set 1, and set $m=50$ for the large instances of data set 2. For the following four parameters, we uniformly choose $\omega=5$, $\alpha=1$, $\beta=10$, $H=10n$ as the default settings. {Finally, for parameter $T$ which controls the termination time, we respectively set $T=10n$ and $T=40n$ milliseconds for each instance of data set 1 and data set 2, to ensure that our algorithm elapses no more time than the best (in terms of solution quality) learning algorithm proposed in each reference paper.}


\subsection{Results on data set 1}

\begin{table*}[!htb]
\small
\centering

\label{table:Exps01}
    \begin{tabular}{l|l|lll|lll|lll}
       \toprule[2pt]
       \multirow{2}{*}{Method} & \multirow{2}{*}{Type}  & \multicolumn{3}{c}{TSP20} & \multicolumn{3}{c}{TSP50} & \multicolumn{3}{c}{TSP100}\\
         &  & Length & Gap & Time & Length & Gap & Time & Length & Gap & Time \\
        \hline
        Concorde  &Exact Solver & {3.8303} & {0.0000\%} & {2.31m} & {5.6906} & {0.0000\%} & {13.68m} & {7.7609} & {0.0000\%} & {1.04h}  \\
        Gurobi  & Exact Solver & {3.8302} & {-0.0001\%} & {2.33m} & {5.6905} & {0.0000\%} & {26.20m} & {7.7609} & {0.0000\%} & {3.57h}  \\
        LKH3  & Heuristic & {3.8303} & {0.0000\%} & {20.96m} & {5.6906} & {0.0013\%} & {26.65m} & {7.7611} & {0.0026\%} & {49.96m}  \\
        \hline
         GAT \citep{deudon2018learning}  & RL, S & \textcolor[rgb]{0,0,1}{3.8741} & \textcolor[rgb]{0,0,1}{1.1443\%} &  \textcolor[rgb]{0,0,1}{10.30m} & \textcolor[rgb]{0,0,1}{6.1085} & \textcolor[rgb]{0,0,1}{7.3438\%} &  \textcolor[rgb]{0,0,1}{19.52m} & \textcolor[rgb]{0,0,1}{8.8372} & \textcolor[rgb]{0,0,1}{13.8679\%} &  \textcolor[rgb]{0,0,1}{47.78m}  \\

        GAT \citep{deudon2018learning}  & \tabincell{c}{RL, S,\\ 2OPT} & \textcolor[rgb]{0,0,1}{3.8501} & \textcolor[rgb]{0,0,1}{0.5178\%} & \textcolor[rgb]{0,0,1}{15.62m} & {5.8941} & {3.5759\%} & {27.81m} & {8.2449} & {6.2365\%} & {4.95h}  \\

        GAT (Kool et al. 2018)  & RL, S & \textcolor[rgb]{0,0,1}{3.8322} & \textcolor[rgb]{0,0,1}{0.0501\%} &  \textcolor[rgb]{0,0,1}{16.47m}  & {5.7185} & {0.4912\%} &  {22.85m} & {7.9735} & {2.7391\%} & {1.23h}  \\

        GAT (Kool et al. 2018)  & RL, G & {3.8413} & {0.2867\%} & {6.03s} & {5.7849} & {1.6568\%} & {34.92s} & {8.1008} & {4.3791\%} & {1.83m} \\

        GAT (Kool et al. 2018)  & RL, BS & \textcolor[rgb]{0,0,1}{3.8304} & \textcolor[rgb]{0,0,1}{0.0022\%} & \textcolor[rgb]{0,0,1}{15.01m} & {5.7070} & {0.2892\%} & {25.58m} & {7.9536} & {2.4829\%} & {1.68h}  \\

        GCN (Joshi et al. 2019) & SL, G & {3.8552} & {0.6509\%} &  {{19.41s}} & {5.8932} & {3.5608\%} &  {2.00m} & {8.4128} & {8.3995\%} &  {11.08m}  \\

        GCN (Joshi et al. 2019) & SL, BS & \textcolor[rgb]{0,0,1}{3.8347} & \textcolor[rgb]{0,0,1}{0.1158\%} &  \textcolor[rgb]{0,0,1}{21.35m} & \textcolor[rgb]{0,0,1}{5.7071} & \textcolor[rgb]{0,0,1}{0.2905\%} &  \textcolor[rgb]{0,0,1}{35.13m} & \textcolor[rgb]{0,0,1}{7.8763} & \textcolor[rgb]{0,0,1}{1.4828\%} &  \textcolor[rgb]{0,0,1}{31.80m}  \\

        GCN (Joshi et al. 2019) & SL, BS* & {3.8305} & {0.0075\%} & {22.18m} & {5.6920} & {0.02509\%} & {37.56m} & {7.8719} & {1.4299\%} & {1.20h}  \\

        \hline
        \multirow{2}{*}{Att-GCRN+MCTS(\textbf{Ours})} & \multirow{2}{*}{SL+RL} & \multirow{2}{*}{{3.8303}} & \multirow{2}{*}{\textbf{0.0000\%}} & {23.33s} +  & \multirow{2}{*}{5.6914} & \multirow{2}{*}{\textbf{0.0145\%}} & {2.59m} +  & \multirow{2}{*}{{7.7638}} & \multirow{2}{*}{\textbf{0.0370\%}} & {3.94m} +   \\
        & & & & 1.25m & & & 5.33m & & & 10.62m\\

        \bottomrule[2pt]
    \end{tabular}
    \caption{Results of Att-GCRN+MCTS w.r.t. existing baselines, tested on 10,000 instances respectively with $n$=20, 50 and 100.}
    \label{table:Exps01}
\end{table*}

Table \ref{table:Exps01} presents the results obtained by our algorithm (Att-GCRN+MCTS) on data set 1, with respect to the existing baselines. Respectively, the first three lines list two exact solvers, i.e., Concorde \citep{applegate2006concorde} \footnote{Downloaded from https://github.com/jvkersch/pyconcorde} and Gurobi \footnote{See https://www.gurobi.com}, as well as one strong heuristic LKH3 \citep{helsgaun2017extension}. The following eight lines are all learning based algorithms which combine traditional operations for post-optimization. There are also several End-to-End ML models in the literature, but they all produce very poor results, thus being omitted here. For the columns, column 1 indicates the methods, while column 2 indicates the type of each algorithm. Columns 3-5 respectively give the average tour length, average gap in percentage w.r.t. Concorde, and the total clock time used by each algorithm on all the 10,000 instances with $n=20$. To ensure fair comparisons, for some learning baselines, the original parameters (such as the width of beam search) are adapted to prolong the total running time. These adapted results are indicated in \textcolor[rgb]{0,0,1}{blue color} in the table. For our method (last line), the time is divided into two parts, i.e., the time for building heat maps plus the time for running MCTS. Columns 6-8, 9-11 respectively give the same information on the instances with $n=50$ and $100$.

As shown in Table \ref{table:Exps01}, the three non-learning algorithms obtain good results on all the test instances, while the existing learning based algorithms all struggle to match optimality on the instances with $n=100$. Compared to these baselines, our algorithm performs quite well, which succeeds in matching the ground-truth solutions (reported by Concorde) on most of these instances, corresponding to an average gap of $\textbf{0.0000\%}$, $\textbf{0.0145\%}$, $\textbf{0.0370\%}$ respectively on the instances with $n=20, 50, 100$. The total runtime of our method remains competitive with respect to all the learning baselines only except two (with greedy heuristics), which are deterministic thus the results cannot be improved by prolonging the runtime.

\subsection{Results on data set 2}

\begin{table*}[!htb]
\small
\centering

\label{table:Exps02}
    \begin{tabular}{l|l|lll|lll|lll}
       \toprule[2pt]
       \multirow{2}{*}{Method} & \multirow{2}{*}{Type}  & \multicolumn{3}{c}{TSP200} & \multicolumn{3}{c}{TSP500} & \multicolumn{3}{c}{TSP1000}\\
         &  & Length & Gap & Time & Length & Gap & Time & Length & Gap & Time \\
        \hline
                Concorde &Solver & {10.7191} & {0.0000\%} & {3.44m} & {16.5458} & {0.0000\%} & {37.66m} & {23.1182} & {0.0000\%} & {6.65h} \\

        Gurobi &Solver & {10.7036} & {-0.1446\%} & {40.49m} & {16.5171} & {-0.1733\%} & {45.63h} & {-} & {-} & {-}  \\

        LKH3 & Heuristic & {10.7195} & {0.0040\%} & {2.01m} & {16.5463} & {0.0029\%} & {11.41m} & {23.1190} & {0.0036\%} & {38.09m}  \\

        \hline
        GAT \citep{deudon2018learning}   & RL, S & \textcolor[rgb]{0,0,1}{13.1746} & \textcolor[rgb]{0,0,1}{22.9079\%} &  \textcolor[rgb]{0,0,1}{4.84m} & \textcolor[rgb]{0,0,1}{28.6291} & \textcolor[rgb]{0,0,1}{73.0293\%} &  \textcolor[rgb]{0,0,1}{20.18m} & \textcolor[rgb]{0,0,1}{50.3018} & \textcolor[rgb]{0,0,1}{117.5860\%} &  \textcolor[rgb]{0,0,1}{37.07m}  \\

        GAT \citep{deudon2018learning} & \tabincell{c}{RL, S,\\ 2OPT} & {11.6104} & {8.3159\%} & {9.59m} & {23.7546} & {43.5687\%} & {57.76m} & {47.7291} & {106.4575\%} & {5.39h}  \\

        GAT (Kool et al. 2018)  & RL, S & \textcolor[rgb]{0,0,1}{11.4497} & \textcolor[rgb]{0,0,1}{6.8160\%} &  \textcolor[rgb]{0,0,1}{4.49m} & {22.6409} & {36.8382\%} & {15.64m} & \textcolor[rgb]{0,0,1}{42.8036} & \textcolor[rgb]{0,0,1}{85.1519\%} &  \textcolor[rgb]{0,0,1}{63.97m}  \\

        GAT (Kool et al. 2018)  & RL, G & {11.6096} & {8.3081\%} & {5.03s} & {20.0188} & {20.9902\%} & {1.51m} & {31.1526} & {34.7539\%} & {3.18m}  \\

        GAT (Kool et al. 2018)  & RL, BS & {11.3769} & {6.1364\%} & {5.77m} & {19.5283} & {18.0257\%} & {21.99m} & {29.9048} & {29.2359\%} & {1.64h}  \\

        GCN (Joshi et al. 2019) & SL, G & {17.0141} & {58.7272\%} &  {59.11s} & {29.7173} & {79.6063\%} & {6.67m} & {48.6151} & {110.2900\%} &  {28.52m}  \\

        GCN (Joshi et al. 2019) & SL, BS & \textcolor[rgb]{0,0,1}{16.1878} & \textcolor[rgb]{0,0,1}{51.0185\%} &  \textcolor[rgb]{0,0,1}{4.63m} & \textcolor[rgb]{0,0,1}{30.3702} & \textcolor[rgb]{0,0,1}{83.5523\%} &  \textcolor[rgb]{0,0,1}{38.02m} & {51.2593} & {121.7278\%} &  {51.67m}  \\

        GCN (Joshi et al. 2019) & SL, BS* & {16.2081} & {51.2079\%} & {3.97m} & {30.4258} & {83.8883\%} & {30.62m} & {51.0992} & {121.0357\%} & {3.23h}  \\

        \hline
        \multirow{2}{*}{Att-GCN+MCTS (\textbf{Ours})}  & \multirow{2}{*}{SL+RL} & \multirow{2}{*}{10.8139} & \multirow{2}{*}{\textbf{0.8844\%}} & {20.62s} + & \multirow{2}{*}{16.9655} & \multirow{2}{*}{\textbf{2.5365\%}} & {31.17s} + & \multirow{2}{*}{23.8634} & \multirow{2}{*}{\textbf{3.2238\%}} & {43.94s} +  \\
        & & & & 2.15m & & & 5.39m & & & 11.74m \\
        \bottomrule[2pt]
    \end{tabular}
    \caption{Results of Att-GCRN+MCTS w.r.t. existing baselines, tested on 128 instances respectively with $n$=200, 500 and 1000.}
    \label{table:Exps02}
\end{table*}

At first, we summarize in Table \ref{table:Exps02} the results obtained on the 384 instances with $n=200, 500, 1000$. Concorde and LKH3 still perform well on these instances, while Gurobi performs well on the instances with $n$=200 and 500, but fails to terminate within reasonable time on the instances with 1000 cities. For the learning baselines, they all produce results far away from optimality, especially on the instances with 1000 vertices. By contrast, our method is able to obtain, within short time, results very close to optimality (corresponding to a gap of $\textbf{0.8844\%}$, $\textbf{2.5365\%}$ and $\textbf{3.2238\%}$ respectively on the instances with $n=200$, $500$ and $1000$), clearly outperforming the existing learning baselines.

Furthermore, we evaluate the performance of Att-GCN+MCTS on the 16 largest instances with 10,000 vertices. On these large instances, several baseline algorithms face a big challenge. For example, the three learning based algorithms proposed in \citep{joshi2019efficient} all fail due to memory exception (tested on the same platform as previously described), while the two exact solvers (Concorde and Gurobi) as well as the two GAT models in \citep{deudon2018learning} all fail due to time exception (up to five hours is allowed for each instance). Therefore, we exclude these seven baseline algorithms, and just compare our Att-GCRN+MCTS algorithm with the remaining three learning based algorithms \citep{kool2018attention}, all evaluated on one GTX 1080 Ti GPU. The results produced by LKH3 (evaluated on one Intel(R) Xeon(R) Gold 5118 CPU @ 2.30GHz) are listed for indicative purpose. As shown in Table \ref{table:largeinstance}, Att-GCRN+MCTS is able to produce solutions close to LKH3, corresponding to a small average gap of \textbf{4.3902$\%$}. By contrast, the three learning based algorithms correspond to a huge average gap of 501.2737$\%$, 97.3932$\%$, 80.2802$\%$ respectively. The runtime of our algorithm remains reasonable (shorter than the best one of the three baselines).



\begin{table}[!htb]
\tiny
\centering

\label{table:largeinstance}
    \begin{tabular}{l|l|lll}
        \toprule[2pt]
       \multirow{2}{*}{Method} & \multirow{2}{*}{Type}  & \multicolumn{3}{c}{TSP10000}\\
         &  & Length & Gap (vs. LKH3) & Time \\
        \hline
        LKH3 & Heuristic & 71.7778 & - & 8.8h \\
        \hline

        GAT (Kool et al. 2018) & RL, S & {431.5812} & {501.2737\%} & {12.63m}  \\
        GAT (Kool et al. 2018) & RL, G & {141.6846} & {97.3932\%} & {5.99m}    \\
        GAT (Kool et al. 2018) & RL, BS & {129.4012} & {80.2802\%} & {1.81h}   \\

        \hline
        \multirow{2}{*}{Att-GCN+MCTS (\textbf{Ours})}  & \multirow{2}{*}{SL+RL} & \multirow{2}{*}{74.9290} & \multirow{2}{*}{\textbf{4.3902\%}} & {4.16m} + \\
        & & & & 1.69h \\

        \bottomrule[2pt]
    \end{tabular}
    \caption{Performance of Att-GCRN+MCTS w.r.t. four baselines, tested on 16 TSP instances with 10,000 vertices.}
    \label{table:largeinstance}
\end{table}


Additionally, we would like to mention two MCTS based TSP algorithms, i.e., \cite{shimomura2016} and \cite{xing2020a}. The source codes of these two papers are both not publicly available, thus we cannot evaluate them uniformly on the same platform to make strictly fair comparisons. In \cite{shimomura2016}, the authors did not report instance-per-instance results, thus it seems impossible for us to make direct comparisons with this method. In \cite{xing2020a}, on the test instances with 20, 50, 100, 200, 500, 1000 cities, the authors respectively claimed an average gap of 0.01\%, 0.20\%, 1.04\%, 1.91\%, 4.37\%, 4.48\% with respect to optimality (all worse than ours), while the time elapsed on each instance was much longer than ours. Roughly speaking, compared to this recent MCTS algorithm, our algorithm is able to produce overall better results within reasonable time (although evaluated on different platforms).


%

\subsection{Ablation study about heat map}

{To emphasize the importance of the heat map, for each instance, we assign an equal probability to each edge, and rerun the MCTS algorithm alone to search solutions. The results are summarized in Table \ref{table:heatmap}, where the left part lists the results obtained by the original Att-GCRN+MCTS algorithm, and the right part lists the results obtained by MCTS alone (without heat map). Clearly, after disabling the heat map, the performance of the algorithm decreases drastically, corresponding to a huge gap with respect to optimality on each data set. For comparison, the original Att-GCRN+MCTS algorithm produces results very close to optimality on each data set. These comparisons clearly certificate the value of the method for identifying promising candidate edges.}

\begin{table}[!htb]
\tiny
\centering

\label{table:heatmap}
    \begin{tabular}{l|lll|lll}
        \toprule[2pt]
       \multirow{2}{*}{Instance} &  \multicolumn{3}{c}{Att-GCRN+MCTS} &  \multicolumn{3}{c}{MCTS (without heat map)} \\
         &  Length & Opt. Gap. & Time &  Length & Opt. Gap. & Time \\
        \hline
        \multirow{2}{*}{TSP20} & \multirow{2}{*}{{3.8303}} & \multirow{2}{*}{{0.0000\%}} & {23.33s}+ & \multirow{2}{*}{7.9934} & \multirow{2}{*}{108.6886\%} & \multirow{2}{*}{{1.16m}} \\
        & & & {1.25m} & & &  \\
        \hline
        \multirow{2}{*}{TSP50} & \multirow{2}{*}{5.6914} & \multirow{2}{*}{0.0145\%} & {2.59m}+ & \multirow{2}{*}{22.0878} & \multirow{2}{*}{288.1454\%} & \multirow{2}{*}{5.19m} \\
        & & & 5.33m & & & \\
        \hline
        \multirow{2}{*}{TSP100} & \multirow{2}{*}{{7.7638}} & \multirow{2}{*}{0.0370\%} & {3.94m}+ & \multirow{2}{*}{46.4342} & \multirow{2}{*}{498.3095\%} & \multirow{2}{*}{10.37m} \\
        & & & 10.62m & & & \\
        \hline
        \multirow{2}{*}{TSP200} & \multirow{2}{*}{10.8139} & \multirow{2}{*}{0.8844\%} & {20.63s}+ & \multirow{2}{*}{96.4989} & \multirow{2}{*}{800.2519\%} & \multirow{2}{*}{2.10m} \\
        & & & 2.15m & & & \\
        \hline
        \multirow{2}{*}{TSP500} & \multirow{2}{*}{16.9655} & \multirow{2}{*}{2.5365\%} & {31.73s}+ & \multirow{2}{*}{247.8806} & \multirow{2}{*}{1398.1482\%} & \multirow{2}{*}{4.81m} \\
        & & & 5.39m & & & \\
        \hline
        \multirow{2}{*}{TSP1000} & \multirow{2}{*}{23.8634} & \multirow{2}{*}{3.2238\%} & {43.94s}+ & \multirow{2}{*}{502.5129} & \multirow{2}{*}{2073.6679\%} & \multirow{2}{*}{10.43m} \\
        & & & 11.74m & & & \\
        \hline
        \multirow{2}{*}{TSP10000} & \multirow{2}{*}{74.9290} & \multirow{2}{*}{4.3902\%} & {4.16m}+ & \multirow{2}{*}{1000.0237} & \multirow{2}{*}{1293.2206\%} & \multirow{2}{*}{1.56h} \\
        & & & 1.69h & & & \\
        \bottomrule[2pt]
    \end{tabular}
    \caption{Ablation study about the heat map.}
    \label{table:heatmap}
\end{table}

%
%
%

\newpage
\section{Conclusions}
\label{Conclusion}

Supervised learning based techniques are useful for discovering common patterns, but require a large amount of training data, being difficult to generalize to large-scale TSP instances. This research shows that, it is possible to train a small-scale model in supervised manner, and smoothly generalize it to tackle large TSP instances, by applying a series of techniques such as graph sampling, graph converting and heat maps merging. This method can inherit the advantages of supervised learning, and avoid repetitively training models of different sizes. Experimental results confirmed that, this method is able to develop highly competitive learning based TSP algorithm, and significantly improve the generalization ability of the pre-trained model. In the future, we will try to solve larger TSP instances or {non-Euclidean TSP instances}, and extend the method to other challenging optimization problems.

\section{Acknowledgements}
We would like to thank the anonymous reviewers for their insightful comments that helped to considerably improve the paper. This paper was supported in part by the Shenzhen Science and Technology Innovation Commission under grant JCYJ20180508162601910, the National Key R\&D Program of China under grant 2020YFB1313300, and the Funding from the Shenzhen Institute of Artificial Intelligence and Robotics for Society under grant 2019-INT003. Jia-Ming Xin also contributed to this paper.

\begin{thebibliography}{10}

\bibitem{applegate2006concorde}
David Applegate, Ribert Bixby, Vasek Chvatal, and William Cook.
\newblock Concorde tsp solver. http://www.math.uwaterloo.ca/tsp/concorde, 2006.

\bibitem{applegate2003chained}
David Applegate, William Cook, and Andr{\'e} Rohe.
\newblock Chained lin-kernighan for large traveling salesman problems.
\newblock {\em INFORMS Journal on Computing}, 15(1):82--92, 2003.

\bibitem{applegate2009certification}
David~L Applegate, Robert~E Bixby, Va{\v{s}}ek Chv{\'a}tal, William Cook,
  Daniel~G Espinoza, Marcos Goycoolea, and Keld Helsgaun.
\newblock Certification of an optimal tsp tour through 85,900 cities.
\newblock {\em Operations Research Letters}, 37(1):11--15, 2009.

\bibitem{bello2016neural}
Irwan Bello, Hieu Pham, Quoc~V Le, Mohammad Norouzi, and Samy Bengio.
\newblock Neural combinatorial optimization with reinforcement learning.
\newblock In {\em Proceeding of the International Conference on Learning
  Representations~(ICLR)}, 2017.

\bibitem{bengio2018machine}
Yoshua Bengio, Andrea Lodi, and Antoine Prouvost.
\newblock Machine learning for combinatorial optimization: a methodological
  tour d'horizon.
\newblock {\em arXiv preprint arXiv:1811.06128}, 2018.

\bibitem{browne2012survey}
Cameron~B Browne, Edward Powley, Daniel Whitehouse, Simon~M Lucas, Peter~I
  Cowling, Philipp Rohlfshagen, Stephen Tavener, Diego Perez, Spyridon
  Samothrakis, and Simon Colton.
\newblock A survey of monte carlo tree search methods.
\newblock {\em IEEE Transactions on Computational Intelligence and AI in
  games}, 4(1):1--43, 2012.

\bibitem{chen2019learning}
Xinyun Chen and Yuandong Tian.
\newblock Learning to perform local rewriting for combinatorial optimization.
\newblock In {\em Advances in Neural Information Processing Systems}, pages
  6278--6289, 2019.

\bibitem{coulom2006efficient}
R{\'e}mi Coulom.
\newblock Efficient selectivity and backup operators in monte-carlo tree
  search.
\newblock In {\em International conference on computers and games}, pages
  72--83. Springer, 2006.

\bibitem{deudon2018learning}
Michel Deudon, Pierre Cournut, Alexandre Lacoste, Yossiri Adulyasak, and
  Louis-Martin Rousseau.
\newblock Learning heuristics for the tsp by policy gradient.
\newblock In {\em International Conference on the Integration of Constraint
  Programming, Artificial Intelligence, and Operations Research}, pages
  170--181. Springer, 2018.

\bibitem{dudani1976distance}
Sahibsingh~A Dudani.
\newblock The distance-weighted k-nearest-neighbor rule.
\newblock {\em IEEE Transactions on Systems, Man, and Cybernetics},
  (4):325--327, 1976.

\bibitem{emami2018learning}
Patrick Emami and Sanjay Ranka.
\newblock Learning permutations with sinkhorn policy gradient.
\newblock {\em arXiv preprint arXiv:1805.07010}, 2018.

\bibitem{google2016ortools}
Google.
\newblock Or-tools, google optimization tools.
\newblock {\em https://developers.google.com/optimization/routing}, 2016.

\bibitem{guo2019solving}
Tiande Guo, Congying Han, Siqi Tang, and Man Ding.
\newblock Solving combinatorial problems with machine learning methods.
\newblock In {\em Nonlinear Combinatorial Optimization}, pages 207--229.
  Springer, 2019.

\bibitem{gurobi2015gurobi}
Incorporate Gurobi~Optimization.
\newblock Gurobi optimizer reference manual.
\newblock {\em URL http://www.gurobi.com}, 2015.

\bibitem{helsgaun2000effective}
Keld Helsgaun.
\newblock An effective implementation of the lin--kernighan traveling salesman
  heuristic.
\newblock {\em European Journal of Operational Research}, 126(1):106--130,
  2000.

\bibitem{helsgaun2009general}
Keld Helsgaun.
\newblock General k-opt submoves for the lin--kernighan tsp heuristic.
\newblock {\em Mathematical Programming Computation}, 1(2-3):119--163, 2009.

\bibitem{helsgaun2017extension}
Keld Helsgaun.
\newblock An extension of the lin-kernighan-helsgaun tsp solver for constrained
  traveling salesman and vehicle routing problems.
\newblock {\em Roskilde: Roskilde University}, 2017.

\bibitem{hopfield1985neural}
John~J Hopfield and David~W Tank.
\newblock Neural computation of decisions in optimization problems.
\newblock {\em Biological cybernetics}, 52(3):141--152, 1985.

\bibitem{joshi2019efficient}
Chaitanya~K Joshi, Thomas Laurent, and Xavier Bresson.
\newblock An efficient graph convolutional network technique for the travelling
  salesman problem.
\newblock {\em arXiv preprint arXiv:1906.01227}, 2019.

\bibitem{kaempfer2018learning}
Yoav Kaempfer and Lior Wolf.
\newblock Learning the multiple traveling salesmen problem with permutation
  invariant pooling networks.
\newblock In {\em Proceeding of the International Conference on Learning
  Representations~(ICLR)}, 2019.

\bibitem{khalil2017learning}
Elias Khalil, Hanjun Dai, Yuyu Zhang, Bistra Dilkina, and Le~Song.
\newblock Learning combinatorial optimization algorithms over graphs.
\newblock In {\em Advances in Neural Information Processing Systems~(NeurIPS)},
  pages 6348--6358, 2017.

\bibitem{kool2018attention}
Wouter Kool, Herke van Hoof, and Max Welling.
\newblock Attention, learn to solve routing problems!
\newblock In {\em International Conference on Learning Representations~(ICLR)},
  2019.

\bibitem{lemos2019graph}
Henrique Lemos, Marcelo Prates, Pedro Avelar, and Luis Lamb.
\newblock Graph colouring meets deep learning: Effective graph neural network
  models for combinatorial problems.
\newblock {\em arXiv preprint arXiv:1903.04598}, 2019.

\bibitem{li2018combinatorial}
Zhuwen Li, Qifeng Chen, and Vladlen Koltun.
\newblock Combinatorial optimization with graph convolutional networks and
  guided tree search.
\newblock In {\em Advances in Neural Information Processing Systems~(NeurIPS)},
  pages 539--548, 2018.

\bibitem{lin1973effective}
Shen Lin and Brian~W Kernighan.
\newblock An effective heuristic algorithm for the traveling-salesman problem.
\newblock {\em Operations research}, 21(2):498--516, 1973.

\bibitem{lu2020a}
Hao Lu, Xingwen Zhang, and Shuang Yang.
\newblock A learning-based iterative method for solving vehicle routing
  problems.
\newblock In {\em International Conference on Learning Representations~(ICLR)},
  2020.

\bibitem{mladenovic1997variable}
Nenad Mladenovi{\'c} and Pierre Hansen.
\newblock Variable neighborhood search.
\newblock {\em Computers \& operations research}, 24(11):1097--1100, 1997.

\bibitem{nazari2018reinforcement}
Mohammadreza Nazari, Afshin Oroojlooy, Lawrence Snyder, and Martin Tak{\'a}c.
\newblock Reinforcement learning for solving the vehicle routing problem.
\newblock In {\em Advances in Neural Information Processing Systems~(NeurIPS)},
  pages 9839--9849, 2018.

\bibitem{nowak2017note}
Alex Nowak, Soledad Villar, Afonso~S Bandeira, and Joan Bruna.
\newblock A note on learning algorithms for quadratic assignment with graph
  neural networks.
\newblock In {\em Proceeding of the 34$^{th}$ International Conference on
  Machine Learning~(ICML)}, volume 1050, page~22, 2017.

\bibitem{prates2019learning}
Marcelo Prates, Pedro~HC Avelar, Henrique Lemos, Luis~C Lamb, and Moshe~Y
  Vardi.
\newblock Learning to solve np-complete problems: A graph neural network for
  decision tsp.
\newblock In {\em Proceedings of the AAAI Conference on Artificial
  Intelligence~(AAAI)}, volume~33, pages 4731--4738, 2019.

\bibitem{rego2011traveling}
C{\'e}sar Rego, Dorabela Gamboa, Fred Glover, and Colin Osterman.
\newblock Traveling salesman problem heuristics: Leading methods,
  implementations and latest advances.
\newblock {\em European Journal of Operational Research}, 211(3):427--441,
  2011.

\bibitem{reinelt1991tsplib}
Gerhard Reinelt.
\newblock Tsplib—a traveling salesman problem library.
\newblock {\em ORSA journal on computing}, 3(4):376--384, 1991.

\bibitem{selsam2018learning}
Daniel Selsam, Matthew Lamm, Benedikt B{\"u}nz, Percy Liang, Leonardo de~Moura,
  and David~L Dill.
\newblock Learning a sat solver from single-bit supervision.
\newblock {\em arXiv preprint arXiv:1802.03685}, 2018.

\bibitem{shah2019reinforcement}
Devavrat Shah, Qiaomin Xie, and Zhi Xu.
\newblock On reinforcement learning using monte carlo tree search with
  supervised learning: Non-asymptotic analysis.
\newblock {\em arXiv preprint arXiv:1902.05213}, 2019.

\bibitem{shen2018m}
Yelong Shen, Jianshu Chen, Po-Sen Huang, Yuqing Guo, and Jianfeng Gao.
\newblock M-walk: Learning to walk over graphs using monte carlo tree search.
\newblock In {\em Advances in Neural Information Processing Systems~(NeurIPS)},
  pages 6786--6797, 2018.

\bibitem{shimomura2016}
Masato Shimomura and Yasuhiro Takashima.
\newblock Application of monte-carlo tree search to traveling-salesman problem.
\newblock In {\em The 20th Workshop on Synthesis And System Integration of
  Mixed Information technologies~(SASIMI)}, pages 352--356, 2016.

\bibitem{silver2016mastering}
David Silver, Aja Huang, Chris~J Maddison, Arthur Guez, Laurent Sifre, George
  Van Den~Driessche, Julian Schrittwieser, Ioannis Antonoglou, Veda
  Panneershelvam, Marc Lanctot, et~al.
\newblock Mastering the game of go with deep neural networks and tree search.
\newblock {\em Nature}, 529(7587):484, 2016.

\bibitem{silver2017mastering}
David Silver, Julian Schrittwieser, Karen Simonyan, Ioannis Antonoglou, Aja
  Huang, Arthur Guez, Thomas Hubert, Lucas Baker, Matthew Lai, Adrian Bolton,
  et~al.
\newblock Mastering the game of go without human knowledge.
\newblock {\em Nature}, 550(7676):354, 2017.

\bibitem{smith1999neural}
Kate~A Smith.
\newblock Neural networks for combinatorial optimization: a review of more than
  a decade of research.
\newblock {\em INFORMS Journal on Computing}, 11(1):15--34, 1999.

\bibitem{taillard2019popmusic}
{\'E}ric~D Taillard and Keld Helsgaun.
\newblock Popmusic for the travelling salesman problem.
\newblock {\em European Journal of Operational Research}, 272(2):420--429,
  2019.

\bibitem{vinyals2015pointer}
Oriol Vinyals, Meire Fortunato, and Navdeep Jaitly.
\newblock Pointer networks.
\newblock In {\em Advances in Neural Information Processing Systems~(NeurIPS)},
  pages 2692--2700, 2015.

\bibitem{wang2019learning}
Runzhong Wang, Junchi Yan, and Xiaokang Yang.
\newblock Learning combinatorial embedding networks for deep graph matching.
\newblock {\em arXiv preprint arXiv:1904.00597}, 2019.

\bibitem{xing2020a}
Zhihao Xing and Shikui Tu.
\newblock A graph neural network assisted monte carlo tree search approach to
  traveling salesman problem.
\newblock In {\em IEEE Access}, 2020.

\end{thebibliography}

\end{document}